\documentclass[11pt, a4paper]{article}

\usepackage[utf8]{inputenc}
\usepackage{geometry}
\usepackage{authblk}
\usepackage{amsmath, amssymb, amsfonts}
\usepackage{amsrefs}
\usepackage{graphicx}
\usepackage{booktabs}
\usepackage{float}
\usepackage{caption}
\usepackage{authblk}
\usepackage{multirow}
\usepackage{subcaption}
\usepackage{siunitx}
\usepackage{hyperref}

\title{Overcoming the Limits of Finite Difference Method; Physics-Informed Neural Network for Noisy High-Dimensional Heat Diffusion}

\author{
	Shreesh Bhattarai, Harish Chandra Bhandari\\
	\vspace{-1em}
	\texttt{shreeshb51@gmail.com, harish.bhandari@ku.edu.np}\\
	\vspace{0.1em}
	Department of Mathematics, Kathmandu University\\
	Dhulikhel, Nepal
}
\date{}

\begin{document}
	
	\maketitle
	
	% Abstract
	\begin{abstract}
		\noindent High-dimensional transient heat diffusion under noisy boundary conditions exposes a fundamental limitation of classical numerical methods: accuracy degrades catastrophically where physical noise is unavoidable. This paper presents a Physics-Informed Neural Network (PINN) framework as a systematic solution to this problem across one, two, and three spatial dimensions, establishing clear operational regimes that redefine solver selection in noisy thermal systems. Under 20\% boundary noise in 3D, PINN sustains approximately 91\% accuracy while Finite Difference Method (FDM) collapses to 36\%, a clear decisive advantage. This is further confirmed in a physical copper thermal system, where PINN reduces boundary reconstruction error by 3.3 times under realistic noise conditions. This noise resilience is accompanied by a dimensionality-driven efficiency crossover: PINN requires fewer spacetime nodes than FDM in 3D while achieving superior accuracy, exposing the true cost of classical discretization at scale. These findings reframe solver selection: the decisive axis is not accuracy alone, but noise exposure and dimensionality jointly. When noise and dimensionality are both high, the classical solver paradigm is insufficient; this work provides the foundation to justify PINN as the operational standard in such regimes.
	\end{abstract}
	

	\section{Introduction}
	\label{sec:introduction}
	
	\noindent Transient heat transfer governs many critical engineering processes, including welding, solidification, power electronics, and thermal management in AI chip packages; however, in real-world applications, boundary conditions are never known with complete precision \cite{cai}. Sensor noise, aleatoric uncertainty from stochastic physical processes \cite{bajaj, yang}, and Kapitza thermal resistance at material interfaces \cite{hansaem}, collectively render ill-posed boundary value problems that traditional deterministic solvers struggle to accommodate \cite{cai}, as numerical differentiation tends to amplify measurement errors \cite{pietro}.
	
	\noindent Classical numerical solvers such as FDM and Finite Element Method (FEM) remain the computational standard for heat diffusion \cite{grossmann, batyr}. However, while their mesh-based approaches are well-understood, computationally fast, and accurate under clean conditions \cite{grossmann, batyr}, their reliance on numerical differentiation \cite{batyr} makes them fundamentally susceptible to noise, resulting in error amplification \cite{pietro}. PINN \cite{raissi, lulu} offers a structurally different approach: by embedding governing equations directly into the loss function via automatic differentiation \cite{baydin}, the network simultaneously satisfies the partial differential equation (PDE), boundary condition (BC), and initial condition (IC) \cite{raissi, cai, vitor}. Critically, the physics-based regularization acts as an implicit denoising mechanism \cite{cai, jian, pietro}, a property that becomes particularly important in situations where classical methods are most vulnerable.
	
	\noindent Despite growing interest in both approaches, the literature contains a precise and consequential gap. Existing PINN--FDM comparisons focus predominantly on clean-condition accuracy \cite{savovic, wasif, batyr}. Noise robustness studies exist, targeting fluid mechanics benchmarks such as Burgers and Navier--Stokes equations \cite{jekic}, as well as other nonlinear systems like the Schrödinger equation \cite{pietro}. Heat-equation-specific studies are either restricted to 2D without noise \cite{vitor}, limited to steady-state problems \cite{hansaem}, or concerned with inverse parameter identification rather than forward solution accuracy \cite{wei}. Most critically, no prior work has systematically quantified how noise sensitivity scales with spatial dimensionality for both solver classes in transient heat diffusion providing specific evidence required to justify deploying PINN in noisy high-dimensional thermal systems. This gap motivates the present work, in which we present a PINN framework as a solution for transient heat diffusion across one, two, and three spatial dimensions, under frozen Gaussian boundary noise, benchmarked against analytical Fourier series solutions. The contributions are fourfold:
	\begin{enumerate}
		\item Established a systematic framework with identical frozen Gaussian noise trajectories across all three dimensions and identified a critical accuracy crossover: PINN sustains approximately 91\% accuracy under 20\% boundary noise in 3D while FDM collapses to 36\% accuracy in the same regime.
		
		\item Physical significance is confirmed through a copper thermal system case study, where absolute temperature reconstruction errors are reported in Kelvin (K) under 30 K boundary noise, yielding a 3.3 times reduction in mean boundary error relative to FDM.
		
		\item Demonstrated that PINN requires 2.81 times fewer spacetime nodes than FDM in 3D while achieving superior accuracy, establishing a dimensionality-driven efficiency crossover alongside the accuracy crossover.
		
		\item Conducted exhaustive hyperparameter optimization over 864 PINN configurations, identifying optimal architectures per dimension and demonstrating that adaptive gradient normalization \cite{bischof} outperforms equal loss weighting \cite{pietro}.
	\end{enumerate}
	Together, these results define clear operational regimes: FDM for real-time clean-data applications, PINN when boundary noise or dimensionality is the limiting factor.
	
	\noindent The remainder of this paper is organized as follows: Section 2 reviews related work. Section 3 details the methodology. Section 4 describes the experimental design. Section 5 presents results. Section 6 discusses physical mechanisms and broader implications. Section 7 concludes with practical recommendations.

	\section{Related Work}
	\label{sec:related}
	
	\noindent For well-posed forward problems under clean conditions, classical mesh-based solvers retain clear advantages \cite{karniadakis, jekic}. Sharimbayev et al. \cite{batyr} demonstrated that FDM achieves relative errors approximately two orders of magnitude lower than PINN for 1D and one and a half orders of magnitude lower for 2D Poisson equations, with substantially lower computational cost. Grossmann et al. \cite{grossmann} corroborated this, finding FEM outperforms PINN by factors of 100 to 1000 in wall-clock time for standard boundary value problems. Savović et al. \cite{savovic} reached the same conclusion for the 1D Burgers' equation, noting that Explicit FDM's well-established stability criteria make it preferable for straightforward problems. However, this advantage inverts in high-dimensional settings: Grossmann et al. \cite{grossmann} found that interpolating FEM solutions onto new meshes in 3D is two to three orders of magnitude slower than evaluating a trained PINN.
	
	\noindent PINN demonstrates clear advantages in regimes that stress classical solvers. Khan et al. \cite{wasif} found PINN outperforms FDM for the spring-mass system, with PINN solutions matching the exact solution at multiple time steps while FDM deviations reach as large as 0.80 units under identical conditions. Bueno et al. \cite{vitor} showed that PINN outperform both Forward Time-Centered Space (FTCS) and Alternating-Direction Implicit (ADI) schemes for the 2D heat equation, attributing this to simultaneous solution across all time steps, which eliminates sequential error propagation. Sharimbayev et al. \cite{batyr} further demonstrated PINN superiority for multi-task learning by simultaneously solving differential equations while reconstructing unknown spatially varying coefficients. For fractional-order operators, Pang et al. \cite{guofei} introduced fPINN, combining automatic differentiation with finite difference discretization for fractional derivatives, achieving robust performance under 20\% noise. This signals that noise tolerance is highly problem-specific and that systematic benchmarking across dimensions for a single PDE class remains absent.
	
	\noindent Within heat transfer specifically, PINN studies have established promising results. Oh and Jo \cite{hansaem} demonstrated axis-augmented PINN achieve FEM-comparable accuracy for steady-state heat equations in composite materials with imperfect contact, with formal error bounds tied to the loss functional. However, this restriction to steady-state problems leaves temporal error accumulation, central to transient analysis, unaddressed. Cai et al. \cite{cai} extended PINN to complex industrial scenarios including multiphase Stefan problems with moving interfaces, reporting orders-of-magnitude speedups over OpenFOAM (traditional numerical solver and an industrial-standard tool) for design optimization tasks, and reporting robustness to sparse, corrupted sensor data. Critically, these noise robustness claims lack systematic quantification of noise levels, comparison against classical methods under identical perturbations, or reporting of absolute temperature errors in physical units across spatial dimensions.
	
	\noindent Two studies directly target noisy boundary conditions for the heat equation. Bajaj et al. \cite{bajaj} proposed combining Gaussian process (GP) regression with PINN to preprocess noisy boundary data, testing on the 1D Schrödinger ($\sigma$=0.1), 1D Burgers' ($\sigma$=0.5), and 2D heat ($\sigma$=1.0) equations. GP preprocessing significantly reduced error propagation and prevented convergence to physics-obeying local minima, with a sparse GP variant reducing the $\mathcal{O}(n^3)$ complexity via inducing points. Zhou and Xu \cite{wei} proposed data-guided PINN employing a two-phase training strategy: pretraining on data loss, then physics-constrained fine-tuning, achieving 6 to 9 times faster convergence than standard PINN under white Gaussian noise down to 25 dB. While Zhou and Xu focus on inverse parameter identification, Bajaj et al. demonstrate that GP-smoothing recovers forward solution accuracy; notably, neither study extends to three-dimensional systems. Lu et al. \cite{lulu} attribute PINN noise tolerance in the DeepXDE framework to automatic differentiation operating on the neural surrogate rather than corrupted measurements directly.
	
	\noindent Dedicated noise robustness investigations confirm PINN advantages but reveal important boundaries. Wong et al. \cite{jian} demonstrated that PINN trained on 100\% noisy velocity data (Signal-to-Noise Ratio of 1:1) for the 2D lid-driven cavity problem perform comparably to a vanilla DNN trained on data with ten times less noise, attributing this to Navier-Stokes regularization suppressing overfitting. Esteves \cite{pietro} showed PINN achieves inverse parameter identification with less than 0.2\% relative error from 20\% corrupted 1D Schrödinger measurements, while FDM fails entirely in the same regime due to noise amplification during numerical differentiation. Jekic et al. \cite{jekic} provide the most direct classical-versus-PINN noise comparison, benchmarking against FEM combined with Sequential Least Squares Programming for Burgers' and Navier-Stokes equations: FEM dominates in 1D and 2D, but the performance gap narrows substantially in 3D. Here, a critical failure mode was identified: under extreme noise, PINN exhibits reward-hacking, biasing estimates to minimize loss rather than recover true physics, with adaptive weighting sometimes causing physics loss to diverge rather than regularize.
	
	\noindent Theoretical and methodological advances have begun formalizing these observations. André-Sloan et al. \cite{sebastien} derived that trainable parameters must satisfy $d_N \log d_N \gtrsim N_s \eta^2$ to achieve empirical risk below noise variance, establishing that model capacity must exceed a critical threshold to exploit corrupted data. Kim and Kang \cite{hankyeol} proposed naPINN, embedding an energy-based model to learn non-Gaussian residual distributions, maintaining accuracy under 15\% outlier corruption via trainable reliability gates. Yang et al. \cite{yang} proposed B-PINN, treating weights as probability distributions, using Hamiltonian Monte Carlo for posterior inference to deliver uncertainty quantification under high noise, though at significant computational cost. Xu et al. \cite{xu} proposed the DL-PDE framework that generates autodifferentiated meta-data to stabilize sparse regression, correctly identifying governing equations under 10\% multiplicative noise where finite difference approximations fail beyond 5\%.
	
	\noindent Collectively, these works establish that PINN noise tolerance is real, physically grounded, and theoretically non-trivial. Every noise study examines either a single spatial dimension or a different PDE class, conflating dimensional effects with problem-specific characteristics. No prior work establishes how noise amplification scales from 1D to 3D for transient heat diffusion, nor identifies the critical noise threshold at which classical methods become unreliable while PINN remains viable. The present study addresses this gap directly, providing the first systematic quantification of noise sensitivity scaling across spatial dimensions for transient heat diffusion.

	\section{Methodology}
	\label{sec:methodology}
	
	\noindent This work structures the comparative framework around three components: a shared problem formulation with a common analytical ground truth, parallel FDM and PINN implementations operating under identical conditions, and a frozen noise protocol that ensures boundary perturbations are consistent across all solvers and dimensions. Each design choice is made to isolate solver behavior from experimental confounds.
	
	\subsection{Problem Formulation}
	
	\noindent The transient heat diffusion equation \cite{carslaw} in one, two, and three spatial dimensions is considered here:
	
	\begin{equation}
		\frac{\partial u}{\partial t} = \alpha \nabla^2 u
		\label{eq:heat_equation}
	\end{equation}
	
	\noindent where $u(\mathbf{x}, t)$ represents the dimensionless temperature field, $\alpha = 0.00011644$ m$^2$/s is the thermal diffusivity of copper at 25\unit{\degreeCelsius} \cite{bipm}, and $\nabla^2$ denotes the Laplacian operator.
	
	\noindent The temperature field $u$ is treated as a dimensionless quantity $\theta = (T - T_\infty)/(T_{\text{ref}} - T_\infty)$, where $T_\infty = 0$ denotes the ambient reference and $T_{\text{ref}}$ the characteristic temperature scale \cite{carslaw}. Because the heat equation is linear, the shifted variable $\theta$ satisfies the same governing equation as $u$ \cite{carslaw}. For the physical case study in Section~\ref{sec:case_study}, $T_{\text{ref}} = 150$ K is assigned, consistent with boundary temperatures reported for electronic chip thermal management problems \cite{cai}.
	
	\subsubsection{Initial Condition}
	
	\noindent The initial temperature distribution follows a triangular wave profile \cite{leveque, carslaw}:
	
	\begin{equation}
		u(x, 0) = \begin{cases}
			\frac{2x}{L}, & 0 \leq x \leq \tfrac{L}{2}, \\[2pt]
			\frac{2(L-x)}{L}, & \tfrac{L}{2} < x \leq L,
		\end{cases}
		\label{eq:ic_1d}
	\end{equation}
	
	\noindent For higher dimensions, the initial condition is constructed as a separable product:
	
	\begin{align}
		u(x, y, 0) &= u_x(x, 0) \cdot u_y(y, 0) \quad \text{(2D)} \label{eq:ic_2d} \\
		u(x, y, z, 0) &= u_x(x, 0) \cdot u_y(y, 0) \cdot u_z(z, 0) \quad \text{(3D)} \label{eq:ic_3d}
	\end{align}
	
	\noindent where each factor follows the triangular profile in Eq.~\eqref{eq:ic_1d}.
	
	\subsubsection{Boundary Condition}
	
	\noindent Homogeneous Dirichlet boundary condition is enforced at all domain boundaries \cite{carslaw}:
	
	\begin{equation}
		u(\mathbf{x}, t) = 0, \quad \mathbf{x} \in \partial\Omega, \quad \forall t > 0
		\label{eq:bc}
	\end{equation}
	
	\noindent To simulate realistic thermal systems with measurement uncertainty, additive Gaussian noise to the boundary condition was introduced:
	
	\begin{equation}
		u_{\text{noisy}}(\mathbf{x}, t) = \mathcal{N}(0, \sigma), \quad \mathbf{x} \in \partial\Omega
		\label{eq:bc_noisy}
	\end{equation}
	
	\noindent where $\sigma \in \{0.0, 0.05, 0.10, 0.15, 0.20\}$ controls noise magnitude. A frozen noise protocol \cite{wei} is employed, wherein a single noise trajectory is generated using a fixed random seed and reused across all methods, ensuring identical boundary perturbations for fair comparison.
	
	\noindent Since $u$ represents the dimensionless field $\theta \in [0,1]$, noise magnitude $\sigma$ is expressed relative to $T_{\text{ref}}$; for the physical case study in Section~\ref{sec:case_study}, $\sigma = 0.20$ corresponds to 30 K boundary measurement uncertainty on a 150 K system.
	
	\subsection{Analytical Solution}
	
	\noindent The analytical solution is obtained via separation of variables and Fourier series expansion \cite{evans, carslaw}:
	
	\begin{equation}
		u(x, t) = \sum_{n=1}^{N} b_n \sin\left(\frac{n\pi x}{L}\right) \exp\left(-\alpha \left(\frac{n\pi}{L}\right)^2 t\right)
		\label{eq:analytical_1d}
	\end{equation}
	
	\noindent where the Fourier coefficients for the triangular initial condition is:
	
	\begin{equation}
		b_n = \frac{8}{n^2 \pi^2} \sin\left(\frac{n\pi}{2}\right)
		\label{eq:fourier_coeff}
	\end{equation}
	
	\noindent For multidimensional problems, the solution generalizes through tensor products:
	
	\begin{equation}
		u(x, y, t) = \sum_{n_x=1}^{N} \sum_{n_y=1}^{N} b_{n_x} b_{n_y} \sin\left(\frac{n_x\pi x}{L}\right) \sin\left(\frac{n_y\pi y}{L}\right) \exp\left(-\alpha \left[\left(\frac{n_x\pi}{L}\right)^2 + \left(\frac{n_y\pi}{L}\right)^2\right] t\right)
		\label{eq:analytical_2d}
	\end{equation}
	
	\noindent with analogous extension to three dimensions. $N = 50$ terms are employed, at which point residual boundary error falls below $10^{-17}$ (1D), $10^{-33}$ (2D), and $10^{-49}$ (3D), confirming series convergence to machine precision. The analytical solution serves as ground truth for all error calculations.
	
	\subsection{Finite Difference Method}
	
	\noindent Three finite difference schemes are implemented, all chosen for their well-understood convergence and stability properties: Explicit (forward Euler), Implicit (backward Euler), and Crank-Nicolson \cite{leveque, strikwerda}.
	
	\subsubsection{Explicit Scheme}
	
	\noindent The forward Euler time discretization yields:
	
	\begin{equation}
		u_i^{n+1} = u_i^n + r(u_{i+1}^n - 2u_i^n + u_{i-1}^n)
		\label{eq:fdm_explicit}
	\end{equation}
	
	\noindent where $r = \alpha \Delta t / \Delta x^2$ is the stability parameter. Stability requires $r \leq 0.5$ (1D), $r \leq 0.25$ (2D), and $r \leq 1/6$ (3D) \cite{strikwerda}. The experiment's discretization ($\Delta x = \Delta y = \Delta z \approx 0.0714$ m, $\Delta t = 0.6$ s) yields $r = 0.0137$ (1D), $r = 0.0274$ (2D), and $r = 0.0411$ (3D), all well within stability bounds.
	
	\subsubsection{Implicit Scheme}
	
	\noindent The backward Euler method solves:
	
	\begin{equation}
		(I - r L) u^{n+1} = u^n
		\label{eq:fdm_implicit}
	\end{equation}
	
	\noindent where $L$ represents the discrete Laplacian operator. For 1D, $L$ is tridiagonal with off-diagonal entries $1$ and diagonal entries $-2$. Higher dimensions employ Kronecker products \cite{leveque, strikwerda}:
	
	\begin{align}
		L_{2D} &= I \otimes D_2 + D_2 \otimes I \\
		L_{3D} &= I_z \otimes I_y \otimes D_2^x + I_z \otimes D_2^y \otimes I_x + D_2^z \otimes I_y \otimes I_x
	\end{align}
	
	\noindent where $D_2$ is the 1D second-difference operator and $I$ denotes the identity matrix.
	
	\subsubsection{Crank-Nicolson Scheme}
	
	\noindent The semi-implicit Crank-Nicolson method averages explicit and implicit operators \cite{leveque, strikwerda}:
	
	\begin{equation}
		(I - \tfrac{1}{2}r L) u^{n+1} = (I + \tfrac{1}{2}r L) u^n
		\label{eq:fdm_cn}
	\end{equation}
	
	\noindent This scheme is unconditionally stable and second-order accurate in both space and time \cite{leveque, strikwerda}.
	
	\noindent Every FDM simulations evolve through 100 time steps ($\Delta t = 0.6$ s) with solutions recorded at 60 uniformly spaced evaluation times ($t \in \{1, 2, \ldots, 60\}$ s) on a uniform spatial grid of 15 points per dimension.
	
	\subsection{Physics-Informed Neural Network}
	
	\noindent PINN \cite{raissi} is adopted here as the deep learning baseline, with network architecture, loss function, collocation point smapling, loss weighting strategies, and training protocol detailed in the following subsections.
	
	\subsubsection{Network Architecture}
	
	\noindent This paper employs fully connected feedforward neural network mapping $(d+1)$-dimensional spacetime coordinates to scalar temperature:
	
	\begin{equation}
		u_\theta(\mathbf{x}, t): \mathbb{R}^{d+1} \to \mathbb{R}
		\label{eq:pinn_architecture}
	\end{equation}
	
	\noindent where $\theta$ denotes trainable parameters and $d$ is the spatial dimension. Input coordinates are normalized to $[-1, 1]$:
	
	\begin{equation}
		\tilde{\mathbf{x}} = 2 \frac{\mathbf{x} - \mathbf{x}_{\min}}{\mathbf{x}_{\max} - \mathbf{x}_{\min}} - 1
		\label{eq:input_norm}
	\end{equation}
	
	\noindent Network depth ranges from 2 to 8 hidden layers with widths from 4 to 128 neurons. Activation functions evaluated include $\tanh$, GELU, SiLU, and Mish \cite{jagtap, hankyeol, vitor}. Weight initialization follows Xavier uniform or Kaiming normal schemes \cite{hankyeol, hansaem, jian}.
	
	\subsubsection{Loss Function}
	
	\noindent The PINN loss comprises three components enforcing the PDE, BC, and IC \cite{hankyeol}:
	
	\begin{equation}
		\mathcal{L}_{\text{total}} = w_{\text{PDE}} \mathcal{L}_{\text{PDE}} + w_{\text{BC}} \mathcal{L}_{\text{BC}} + w_{\text{IC}} \mathcal{L}_{\text{IC}}
		\label{eq:loss_total}
	\end{equation}
	
	\noindent The PDE residual loss is computed via automatic differentiation \cite{baydin, hankyeol}:
	
	\begin{equation}
		\mathcal{L}_{\text{PDE}} = \frac{1}{N_{\text{PDE}}} \sum_{i=1}^{N_{\text{PDE}}} \left| \frac{\partial u_\theta}{\partial t} - \alpha \nabla^2 u_\theta \right|^2 \bigg|_{(\mathbf{x}_i, t_i)}
		\label{eq:loss_pde}
	\end{equation}
	
	\noindent Boundary and initial conditions losses enforce constraints via mean squared error \cite{raissi, wei}:
	
	\begin{align}
		\mathcal{L}_{\text{BC}} &= \frac{1}{N_{\text{BC}}} \sum_{j=1}^{N_{\text{BC}}} \left| u_\theta(\mathbf{x}_j, t_j) - u_{\text{BC}}(\mathbf{x}_j, t_j) \right|^2, \quad \mathbf{x}_j \in \partial\Omega \label{eq:loss_bc} \\
		\mathcal{L}_{\text{IC}} &= \frac{1}{N_{\text{IC}}} \sum_{k=1}^{N_{\text{IC}}} \left| u_\theta(\mathbf{x}_k, 0) - u_0(\mathbf{x}_k) \right|^2 \label{eq:loss_ic}
	\end{align}
	
	\noindent where $u_{\text{BC}}$ includes noise perturbations when applicable (Eq.~\eqref{eq:bc_noisy}).
	
	\subsubsection{Collocation Point Sampling}
	
	\noindent Training points are generated using Latin Hypercube Sampling (LHS) \cite{mckay} with dimension-dependent budgets summarized in Table~\ref{tab:collocation}.
	
	\begin{table}[h]
		\centering
		\caption{Collocation point budgets per dimension.}
		\label{tab:collocation}
		\begin{tabular}{lcccc}
			\hline
			Dimension & $N_{\text{PDE}}$ & $N_{\text{BC}}$ & $N_{\text{IC}}$ & Total \\
			\hline
			1D & 3,000  & 1,000  & 2,000  & 6,000   \\
			2D & 30,000 & 10,000 & 20,000 & 60,000  \\
			3D & 60,000 & 20,000 & 40,000 & 120,000 \\
			\hline
		\end{tabular}
	\end{table}
	
	\noindent PDE collocation points are resampled every 100 epochs to improve domain coverage. During L-BFGS \cite{mannel} optimization (Section~\ref{sec:training}), the points remain fixed to ensure convergence stability. Moreover, boundary and initial conditions points remain fixed throughout to maintain temporal consistency with the frozen noise protocol.
	
	\subsubsection{Loss Weighting Strategies}
	
	\noindent Three loss balancing approaches are examined and compared:
	
	\begin{enumerate}
		\item \textbf{Equal weighting:} $w_{\text{PDE}} = w_{\text{BC}} = w_{\text{IC}} = 1$.
		
		\item \textbf{Adaptive gradient normalization \cite{bischof}:} Weights are updated every 20 epochs based on gradient magnitude moving averages:
		\begin{equation}
			w_i(t) = \frac{\bar{G}(t)}{3 \bar{G}_i(t)}, \quad \bar{G}_i(t) = \beta \bar{G}_i(t-1) + (1-\beta) \max(\|\nabla_\theta \mathcal{L}_i\|_2, 10^{-8})
			\label{eq:adaptive_gradnorm}
		\end{equation}
		where $\beta = 0.7$ and weights are clamped to dimension-specific ranges to prevent instability:
		\begin{itemize}
			\item \textbf{1D:} $w_{\text{PDE}} \in [50, 500]$, $w_{\text{BC}} \in [0.1, 10]$, $w_{\text{IC}} \in [0.1, 10]$
			\item \textbf{2D:} $w_{\text{PDE}} \in [50, 500]$, $w_{\text{BC}} \in [0.1, 10]$, $w_{\text{IC}} \in [1, 50]$
			\item \textbf{3D:} $w_{\text{PDE}} \in [25, 750]$, $w_{\text{BC}} \in [0.1, 10]$, $w_{\text{IC}} \in [0.1, 10]$
		\end{itemize}
		
		\item \textbf{Adaptive loss-ratio weighting:} Weights are set proportional to relative loss magnitudes with dimension-dependent PDE boosting \cite{heydari}, with an exponential learning rate scheduler ($\gamma = 0.99$) applied concurrently \cite{pytorch}:
		\begin{equation}
			w_i = \frac{\mathcal{L}_i}{\sum_j \mathcal{L}_j + 10^{-8}} \times \lambda_i
		\end{equation}
		where $\lambda_{\text{PDE}} \in \{1, 10, 100\}$ for 1D, 2D, 3D respectively, and $\lambda_{\text{BC}} = \lambda_{\text{IC}} = 1$.
	\end{enumerate}
	
	\subsubsection{Training Protocol}
	\label{sec:training}
	
	\noindent Models are trained for 6,000 epochs using the Adam optimizer \cite{kingma} with learning rates $\eta \in \{5 \times 10^{-3}, 5 \times 10^{-4}\}$. For hybrid Adam-L-BFGS configurations \cite{jekic}, the optimizer switches to L-BFGS at epoch 5,400 (90\% of total epochs) with strong Wolfe line search \cite{pytorch, mannel}. $L_2$ regularization (weight decay $= 10^{-4}$) is applied exclusively to hybrid configurations. Training is terminated early if total loss exceeds 16.0 after epoch 400, a heuristic threshold selected to prune divergent configurations efficiently, and the best model state is restored upon completion.
	
	\subsection{Hyperparameter Optimization}
	
	\noindent A systematic grid search across 864 configurations were conducted. Table~\ref{tab:hyperparam} summarizes the search space.
	
	\begin{table}[h]
		\centering
		\caption{Hyperparameter search space (864 total configurations).}
		\label{tab:hyperparam}
		\begin{tabular}{ll}
			\hline
			Hyperparameter & Values \\
			\hline
			Network depth & 2, 4, 6, 8 (dimension specific) \\
			Network width & 4, 16, 32, 64, 128 (dimension specific) \\
			Activation function & $\tanh$, GELU, SiLU, Mish \\
			Learning rate & $5 \times 10^{-3}$,\ $5 \times 10^{-4}$ \\
			Optimizer & Adam,\ Adam$\to$L-BFGS \\
			Weight initialization & Xavier uniform,\ Kaiming normal \\
			Loss weighting & Equal,\ Grad. Norm.,\ LR Weighting \\
			\hline
		\end{tabular}
	\end{table}
	
	\noindent The optimal configuration per dimension minimizes relative $L_2$ error evaluated against the analytical solution at the same 15-point spatial grid and 60 temporal snapshots used for FDM validation.
	
	\subsection{Error Metrics}
	
	\noindent The methods are evaluated against the analytical solution using three metrics \cite{hansaem, batyr}:
	
	\begin{align}
		\mathcal{E}_{L_2} &= \sqrt{\sum_{i=1}^{M} (u_i^{\text{pred}} - u_i^{\text{true}})^2} \label{eq:l2_error} \\
		\mathcal{E}_{\infty} &= \max_{i=1,\ldots,M} |u_i^{\text{pred}} - u_i^{\text{true}}| \label{eq:linf_error} \\
		\mathcal{E}_{\text{rel}} &= \frac{\mathcal{E}_{L_2}}{\sqrt{\sum_{i=1}^{M} (u_i^{\text{true}})^2}} \label{eq:rel_l2_error}
	\end{align}
	
	\noindent where $M$ denotes the total number of spatiotemporal evaluation points: 900 (1D), 13,500 (2D), and 202,500 (3D). Accuracy percentage is defined as $(1 - \mathcal{E}_{\text{rel}}) \times 100\%$.
	
	\subsection{Computational Environment}
	
	\noindent The experiments were conducted on an NVIDIA Tesla P100-PCIE-16GB GPU using PyTorch with deterministic CUDA operations enabled \cite{pytorch}. Reproducibility is ensured through fixed random seeds across all stochastic processes: network initialization, collocation sampling, and noise generation. Wall-clock execution time and peak GPU memory consumption are recorded for all configurations.

	\section{Experiments}
	\label{sec:experiments}
	
	\noindent The experimental campaign is structured into four sequential phases, progressing from analytical validation through clean-condition baselines to full noise robustness analysis. All phases share identical spatial grids, temporal snapshots, and frozen noise trajectories to eliminate confounding variables across solver comparisons.
	
	\subsection{Experimental Design}
	
	\subsubsection{Phase 1: Analytical Solution Validation}
	
	\noindent Reference solutions are computed using the Fourier series formulation (Eqs.~\eqref{eq:analytical_1d}--\eqref{eq:analytical_2d}) with $N = 50$ terms, evaluated on a uniform $15^d$ spatial grid at 60 temporal snapshots ($t \in \{1, 2, \ldots, 60\}$ s). These datasets serve as ground truth for all subsequent error calculations. Convergence is verified by comparing the series against the true triangular profile at $t = 0$; $N = 50$ yields a maximum IC approximation error of $8.1 \times 10^{-3}$, consistent with the Gibbs phenomenon inherent to finite Fourier representations of piecewise-linear functions. Boundary residuals at $t > 0$ confirm machine-precision satisfaction of homogeneous Dirichlet conditions ($\leq 10^{-17}$, $10^{-33}$, $10^{-49}$ for 1D, 2D, 3D respectively).
	
	\subsubsection{Phase 2: FDM Baseline}
	
	\noindent The three finite difference schemes are executed under clean BC ($\sigma = 0$) across all dimensions, evolving through 100 time steps ($\Delta t = 0.6$ s) with solutions recorded at the 60 evaluation snapshots matching the analytical ground truth. These runs establish baseline accuracy and computational efficiency benchmarks against which PINN performance is measured.
	
	\subsubsection{Phase 3: PINN Hyperparameter Optimization}
	
	\noindent The 864-configuration grid search is distributed across dimensions as summarized in Table~\ref{tab:hpo_dist}. Each configuration is trained under clean conditions ($\sigma = 0$) for 6,000 epochs following the protocol in Section~\ref{sec:training}. The optimal configuration per dimension is selected by minimum relative $L_2$ error against the analytical ground truth. Early stopping is disabled during this phase to ensure unbiased architectural comparison.
	
	\begin{table}[h]
		\centering
		\caption{Hyperparameter search distribution across dimensions.}
		\label{tab:hpo_dist}
		\begin{tabular}{llcc}
			\hline
			Dimension & Depth & Width & Configurations \\
			\hline
			1D & 2 & 4, 6, 8 & 144 \\
			2D & 4, 6 & 32, 64 & 384 \\
			3D & 6, 8 & 128 & 192 \\
			\hline
			\multicolumn{3}{r}{Total} & 864 \\
			\hline
		\end{tabular}
	\end{table}
	
	\subsubsection{Phase 4: Noise Robustness Analysis}
	
	\noindent Solvers are evaluated under five noise levels $\sigma \in \{0.0, 0.05, 0.10, 0.15, 0.20\}$. Frozen noise trajectories are shared identically across FDM and PINN methods at each dimension-noise pair, ensuring boundary perturbations are controlled rather than confounded. The resulting experiment counts are:
	
	\begin{itemize}
		\item \textbf{FDM:} $3 \text{ schemes} \times 3 \text{ dimensions} \times 5 \text{ noise levels} = 45$ experiments.
		\item \textbf{PINN:} $3 \text{ dimensions} \times 5 \text{ noise levels} = 15$ experiments, each using the optimal configuration from Phase 3.
	\end{itemize}
	
	\noindent Every methods are validated against the same clean analytical solutions, directly measuring each solver's ability to recover the true temperature field from corrupted boundary data.
	
	\subsection{Statistical Validation}
	
	\noindent To evaluate whether adaptive gradient normalization outperforms equal weighting, a supplementary study was conducted training both strategies under 10 independent random seeds at moderate noise ($\sigma = 0.10$) for each dimension, yielding 60 training runs in total. Statistical significance is assessed via the Mann-Whitney U test \cite{whitney}, a non-parametric test requiring no distributional assumptions, with the null hypothesis that equal weighting achieves equivalent or better performance.
	
	\subsection{Computational Efficiency Metrics}
	
	\noindent For all experiments, wall-clock execution time, peak GPU memory consumption, and discretization efficiency (ratio of FDM spacetime nodes to PINN collocation points) are recorded. FDM execution times are averaged over all time-stepping iterations. PINN times include training and validation inference; inference contributes less than 1\% of total runtime and is reported separately for transparency.

	\section{Results}
	\label{sec:results}
	
	\noindent The work present results across four dimensions of analysis: optimal PINN configuration identification, clean-condition accuracy benchmarking, noise robustness characterization, and computational efficiency. Numerical comparisons are made against the analytical Fourier series ground truth.
	
	\subsection{Optimal PINN Configurations}
	
	\noindent Systematic search across 864 configurations identifies distinct optimal architectures per dimension (Table~\ref{tab:best_pinn_configs}). Network complexity scales with dimensionality: 1D requires 2 hidden layers of 8 neurons (105 parameters), 2D requires 6 layers of 64 neurons (21,121 parameters), and 3D requires 6 layers of 128 neurons (83,329 parameters). Notably, all optimal configurations employ $\tanh$ activation and Adam exclusively. Hybrid Adam$\to$L-BFGS switching offered no improvement in any dimension. Adaptive gradient normalization is the superior loss weighting strategy across all dimensions.
	
	\begin{table}[htbp]
		\centering
		\caption{Optimal PINN configurations identified through hyperparameter search.}
		\label{tab:best_pinn_configs}
		\begin{tabular}{lccccccc}
			\hline
			Dim & Depth & Width & Activation & LR & Optimizer & Loss Weight & Init \\
			\hline
			1D & 2 & 8   & tanh & $5 \times 10^{-3}$ & Adam & Adaptive GradNorm & Kaiming \\
			2D & 6 & 64  & tanh & $5 \times 10^{-4}$ & Adam & Adaptive GradNorm & Xavier \\
			3D & 6 & 128 & tanh & $5 \times 10^{-4}$ & Adam & Adaptive GradNorm & Xavier \\
			\hline
		\end{tabular}
	\end{table}
	
	\noindent Training convergence is achieved within 6,000 epochs for all dimensions. The PDE residual consistently attains $\mathcal{O}(10^{-8})$ magnitude, while boundary and initial conditions losses settle at $\mathcal{O}(10^{-5})$ to $\mathcal{O}(10^{-6})$ (Table~\ref{tab:pinn_losses}).
	
	\begin{table}[htbp]
		\centering
		\caption{Training losses for optimal PINN under clean boundary condition.}
		\label{tab:pinn_losses}
		\begin{tabular}{lcccc}
			\hline
			Dimension & $\mathcal{L}_{\text{PDE}}$ & $\mathcal{L}_{\text{BC}}$ & $\mathcal{L}_{\text{IC}}$ & $\mathcal{L}_{\text{total}}$ \\
			\hline
			1D & $2.17 \times 10^{-8}$ & $7.78 \times 10^{-7}$ & $3.11 \times 10^{-5}$ & $3.31 \times 10^{-5}$ \\
			2D & $3.34 \times 10^{-8}$ & $7.07 \times 10^{-6}$ & $2.73 \times 10^{-5}$ & $4.79 \times 10^{-5}$ \\
			3D & $1.81 \times 10^{-8}$ & $1.40 \times 10^{-5}$ & $4.30 \times 10^{-5}$ & $3.94 \times 10^{-5}$ \\
			\hline
		\end{tabular}
	\end{table}
	
\begin{figure}[H]
		\centering
		\includegraphics[width=\textwidth, height=0.45\textheight]{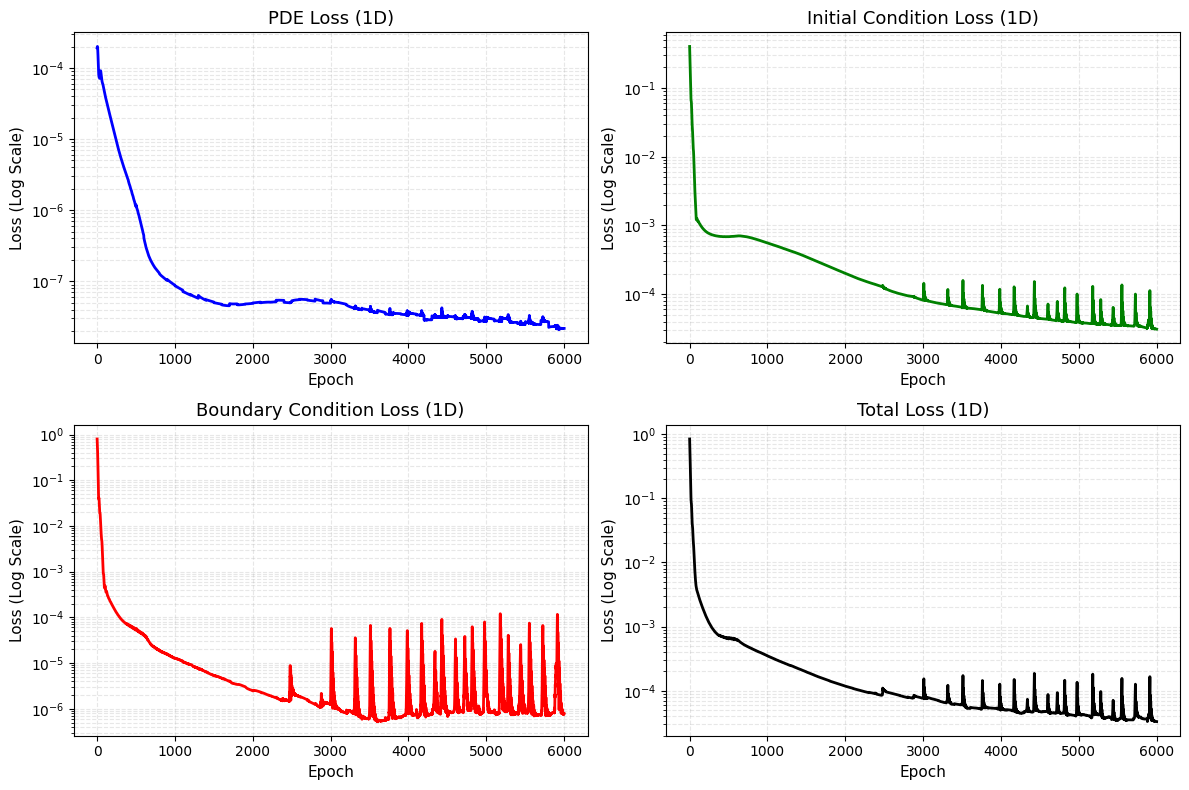}
		\captionof{figure}{Loss convergence curves for optimal PINN configuration in 1D heat equation.}
		\label{fig:1d_loss}
	\end{figure}

\begin{figure}[H]
		\centering
		\includegraphics[width=\textwidth, height=0.45\textheight]{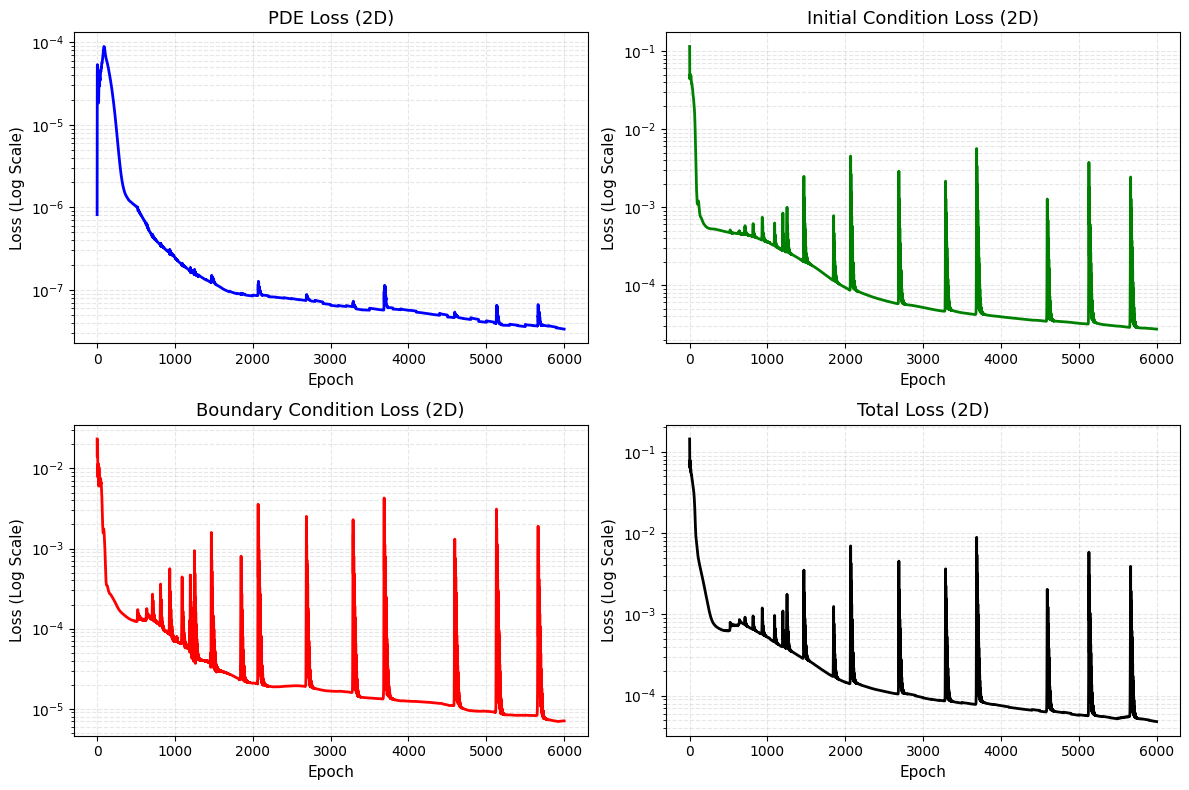}
		\captionof{figure}{Loss convergence curves for optimal PINN configuration in 2D heat equation.}
		\label{fig:2d_loss}
		
		\vspace{1cm}
		
		\includegraphics[width=\textwidth, height=0.45\textheight]{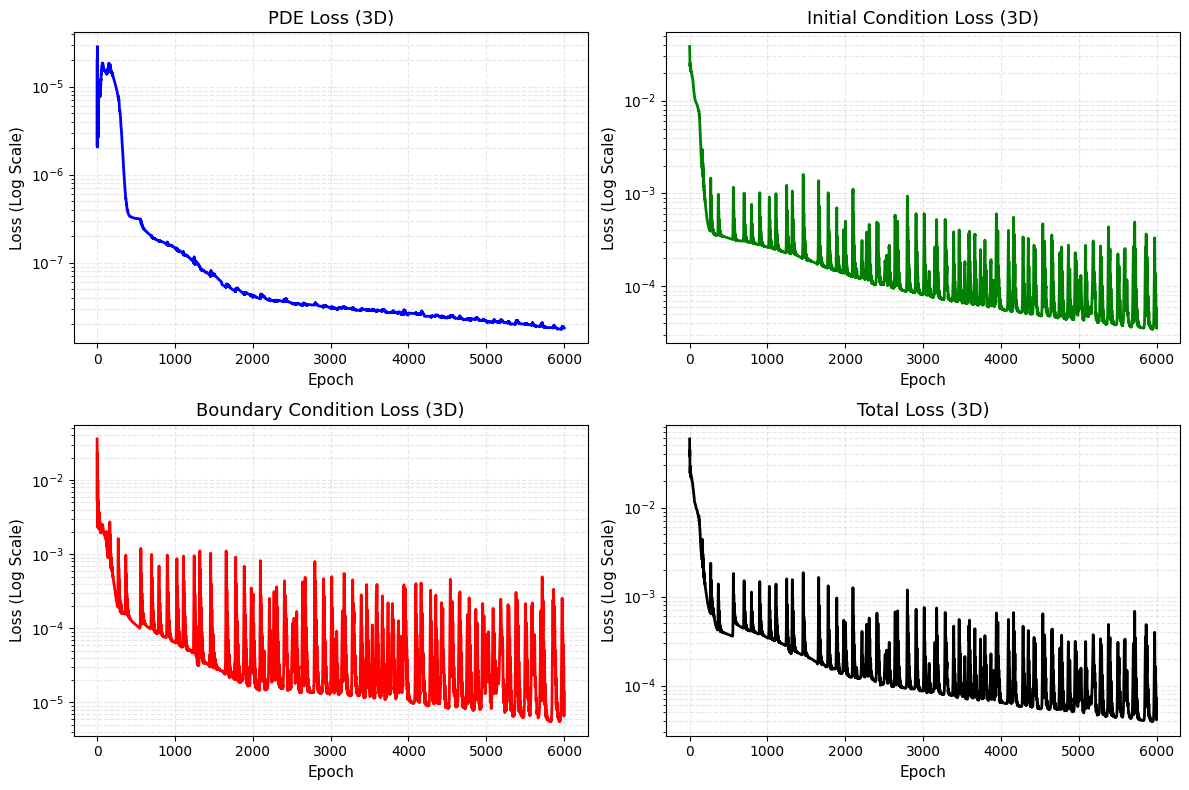}
		\captionof{figure}{Loss convergence curves for optimal PINN configuration in 3D heat equation.}
		\label{fig:3d_loss}
	\end{figure}

	\noindent Figure~\ref{fig:1d_loss} shows the training convergence for the optimal 1D configuration. The PDE loss descends smoothly to $\mathcal{O}(10^{-8})$, while the BC loss and IC loss exhibit periodic spikes consistent with PDE collocation resampling every 100 epochs; when new PDE points are drawn, the loss landscape shifts transiently across all components. The best-model restoration mechanism correctly captures the global minimum rather than these resampling-induced transients. Figure~\ref{fig:2d_loss} illustrates 2D training dynamics: spikes are more frequent and larger than in 1D, reflecting the increased sensitivity of the 2D loss landscape to PDE point redistribution. Figure~\ref{fig:3d_loss} reveals qualitatively different 3D dynamics: all loss components exhibit sustained high-frequency oscillations throughout training, indicating that the 3D landscape is significantly more sensitive to resampling. Despite this, the PDE loss converges smoothly to $\mathcal{O}(10^{-8})$, confirming that physics residuals remain well-controlled even as boundary and initial conditions losses oscillate.
	
	\subsection{Accuracy Comparison Under Clean Conditions}
	
	\noindent Under noise-free boundary condition, PINN achieves superior relative $L_2$ accuracy across all dimensions (Table~\ref{tab:clean_accuracy}): 0.276\% vs.\ 0.751\% (1D), 0.845\% vs.\ 1.212\% (2D), and 1.583\% vs.\ 1.646\% (3D), corresponding to error reductions of 2.72 times, 1.43 times, and 1.04 times over the best FDM scheme. The diminishing margin with dimensionality is expected: as problem complexity increases, both methods converge toward similar discretization-limited accuracy floors under clean conditions. Among FDM schemes, the Explicit method consistently achieves the best accuracy across all dimensions.
	
	\begin{table}[H]
		\centering
		\caption{Accuracy metrics for all methods under clean boundary condition ($\sigma = 0$).}
		\label{tab:clean_accuracy}
		\begin{tabular}{llccccc}
			\hline
			Dim & Method & Rel.\ $L_2$ (\%) & $L_2$ Error & $L_\infty$ Error & Accuracy (\%) & Time (s) \\
			\hline
			\multirow{4}{*}{1D}
			& Explicit FDM & 0.751 & 0.121 & 0.0228 & 99.25 & 0.003 \\
			& Implicit FDM & 0.788 & 0.127 & 0.0234 & 99.21 & 0.016 \\
			& CN FDM & 0.770 & 0.124 & 0.0231 & 99.23 & 0.009 \\
			& \textbf{PINN} & \textbf{0.276} & \textbf{0.044} & \textbf{0.0151} & \textbf{99.72} & 71.49s \\
			\hline
			\multirow{4}{*}{2D}
			& Explicit FDM & 1.212 & 0.406 & 0.0435 & 98.79 & 0.005 \\
			& Implicit FDM & 1.268 & 0.425 & 0.0449 & 98.73 & 0.040 \\
			& CN FDM & 1.240 & 0.416 & 0.0442 & 98.76 & 0.041 \\
			& \textbf{PINN} & \textbf{0.845} & \textbf{0.283} & \textbf{0.0325} & \textbf{99.16} & 220.91s \\
			\hline
			\multirow{4}{*}{3D}
			& Explicit FDM & 1.646 & 1.150 & 0.0622 & 98.35 & 0.011 \\
			& Implicit FDM & 1.723 & 1.204 & 0.0646 & 98.28 & 2.53 \\
			& CN FDM & 1.685 & 1.177 & 0.0635 & 98.32 & 2.62 \\
			& \textbf{PINN} & \textbf{1.583} & \textbf{1.106} & 0.0866 & \textbf{98.42} & 893.44s \\
			\hline
		\end{tabular}
	\end{table}
	
	\noindent A notable exception appears in the 3D $L_\infty$ error, where PINN (0.0866) exceeds FDM (0.0622) despite superior $L_2$ performance. This indicates spatially localized errors in the PINN solution, likely near boundary regions where the triangular profile induces sharp gradients, while global accuracy remains higher. This is discussed further in Section~\ref{sec:discussion}.
	
\begin{figure}[H]
		\centering
		\includegraphics[width=\textwidth]{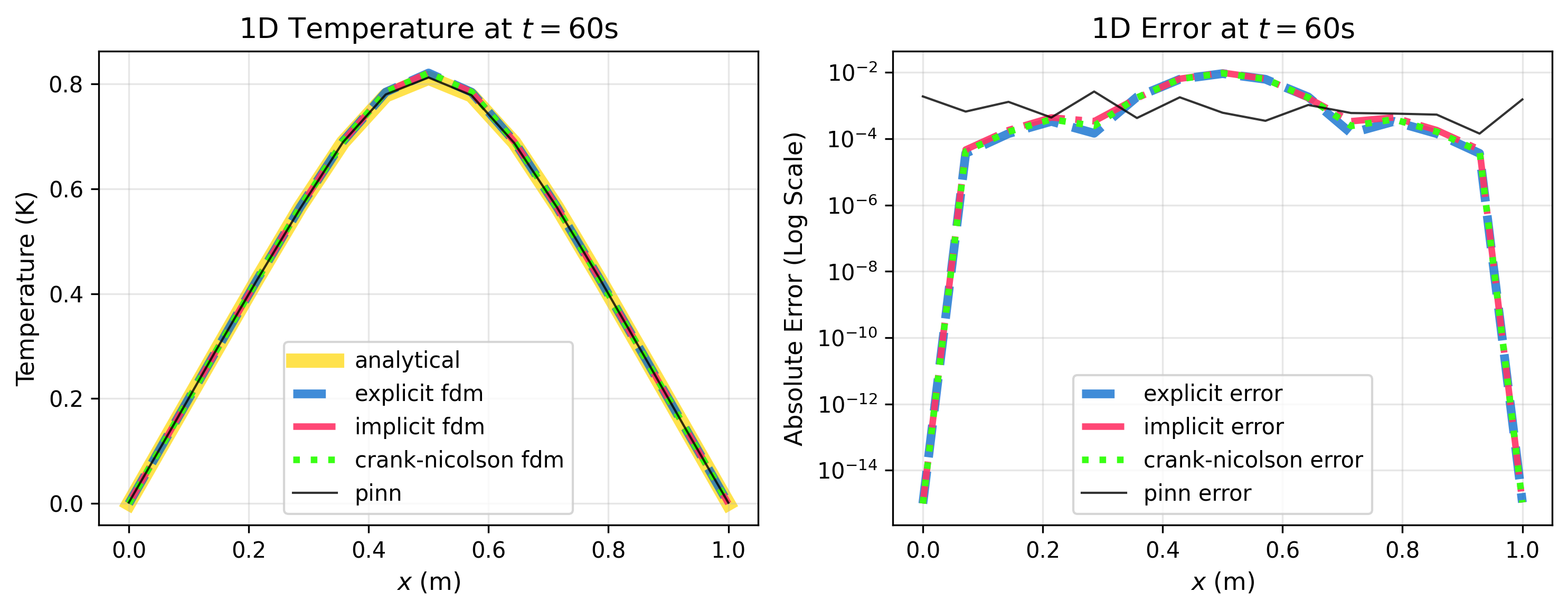}
		\caption{1D heat equation: solution and pointwise error comparison between FDM and PINN under clean boundary condition ($\sigma = 0$).}
		\label{fig:1d_results}
		
		\vspace{1em}
		\includegraphics[width=\textwidth]{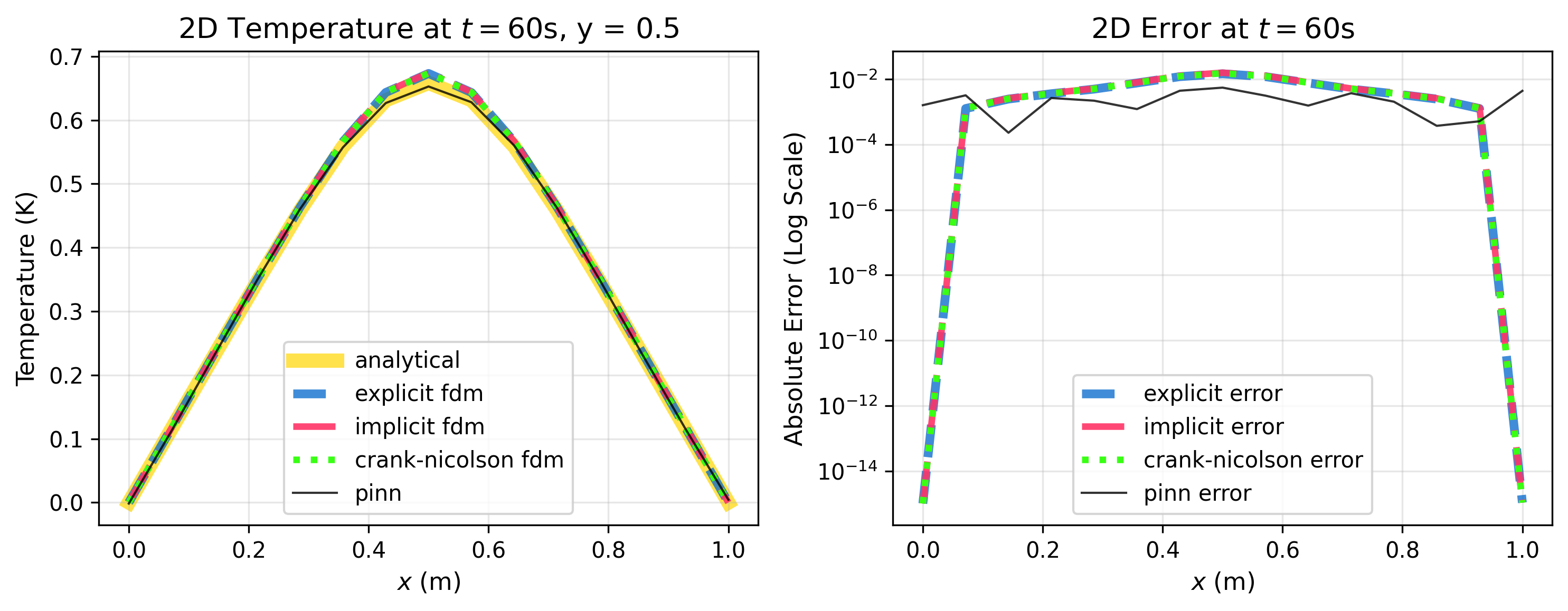}
		\caption{2D heat equation: solution and pointwise error comparison between FDM and PINN under clean boundary condition ($\sigma = 0$).}
		\label{fig:2d_results}
		
		\vspace{1em}
		\includegraphics[width=\textwidth]{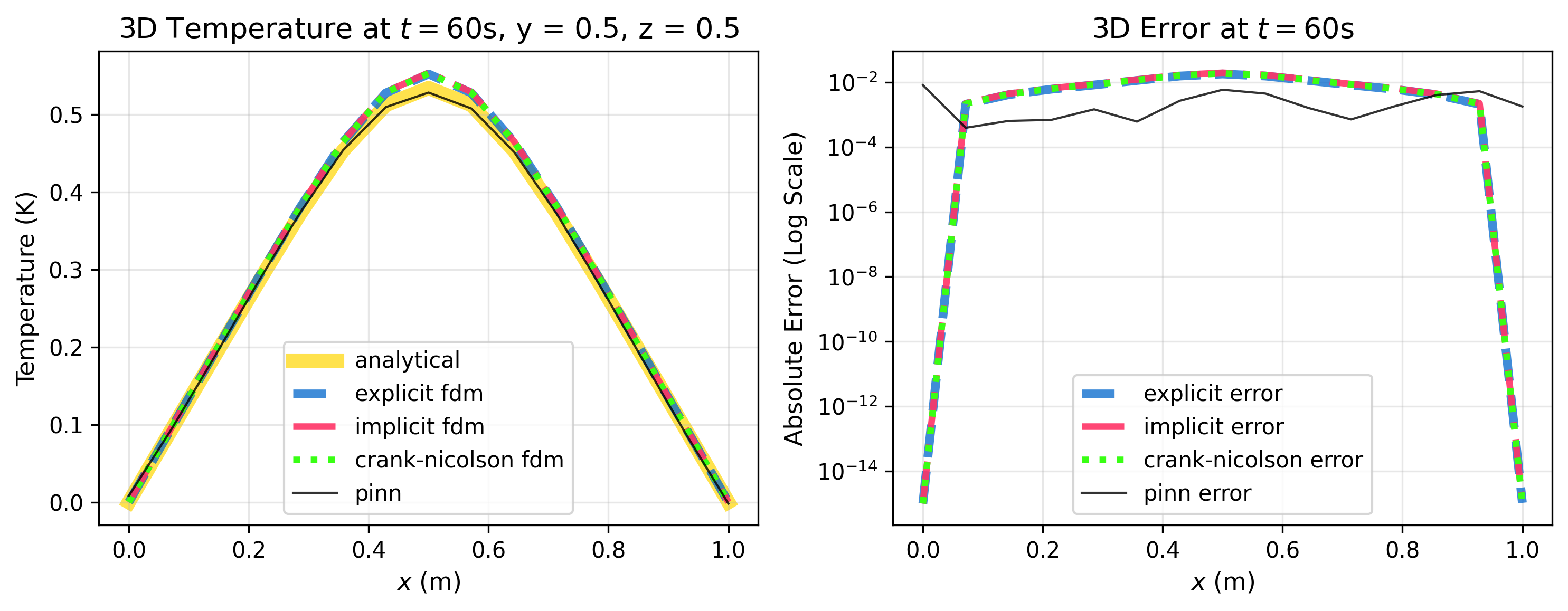}
		\caption{3D heat equation: solution and pointwise error comparison between FDM and PINN under clean boundary condition ($\sigma = 0$).}
		\label{fig:3d_results}
	\end{figure}

	\noindent Figure~\ref{fig:1d_results} confirms close agreement between all methods and the analytical solution at $t = 60$ s. The error panel reveals that all three FDM schemes produce nearly identical absolute errors across the domain, while the PINN error is uniformly lower in the interior but rises near the boundaries, consistent with the soft-constraint treatment of boundary condition. Figure~\ref{fig:2d_results} presents the temperature profile along the centerline ($y = 0.5$ m) at $t = 60$ s. Each method reproduces the analytical solution closely, with FDM schemes overlapping nearly perfectly. The error panel confirms PINN boundary error elevation relative to FDM, mirroring the 1D pattern but with a higher absolute floor consistent with increased problem complexity. Figure~\ref{fig:3d_results} shows the temperature profile along the centerline ($y = 0.5$ m, $z = 0.5$ m) at $t = 60$ s. The PINN solution visibly departs from the analytical curve near the peak, consistent with the elevated $L_\infty$ error reported in Table~\ref{tab:clean_accuracy}. The error panel confirms that while FDM and PINN interior errors are comparable, PINN boundary errors are higher, the localised corner penalty discussed in Section~\ref{sec:discussion}.
	
	\subsection{Noise Robustness Analysis}
	
	\noindent Table~\ref{tab:noise_robustness} reveals fundamentally different degradation regimes between FDM and PINN as noise intensity and dimensionality increase jointly.
	
	\begin{table}[htbp]
		\centering
		\caption{Accuracy (\%) under increasing boundary noise levels across all methods.}
		\label{tab:noise_robustness}
		\begin{tabular}{llccccc}
			\hline
			Dim & Method & $\sigma=0.0$ & $\sigma=0.05$ & $\sigma=0.10$ & $\sigma=0.15$ & $\sigma=0.20$ \\
			\hline
			\multirow{4}{*}{1D}
			& Explicit FDM & 99.25 & 97.06 & 94.27 & 91.44 & 88.61 \\
			& Implicit FDM & 99.21 & 97.05 & 94.26 & 91.44 & 88.61 \\
			& CN FDM & 99.23 & 97.06 & 94.28 & 91.46 & 88.63 \\
			& \textbf{PINN} & \textbf{99.72} & \textbf{99.35} & \textbf{98.89} & \textbf{98.40} & \textbf{98.06} \\
			\hline
			\multirow{4}{*}{2D}
			& Explicit FDM & 98.79 & 92.68 & 85.50 & 78.30 & 71.08 \\
			& Implicit FDM & 98.73 & 92.67 & 85.50 & 78.30 & 71.08 \\
			& CN FDM & 98.76 & 92.69 & 85.53 & 78.34 & 71.14 \\
			& \textbf{PINN} & \textbf{99.16} & \textbf{98.68} & \textbf{97.71} & \textbf{96.68} & \textbf{95.47} \\
			\hline
			\multirow{4}{*}{3D}
			& Explicit FDM & 98.35 & 84.02 & 68.16 & 52.27 & 36.38 \\
			& Implicit FDM & 98.28 & 84.01 & 68.16 & 52.28 & 36.38 \\
			& CN FDM & 98.32 & 84.05 & 68.23 & 52.37 & 36.51 \\
			& \textbf{PINN} & \textbf{98.42} & \textbf{94.37} & \textbf{94.23} & \textbf{93.16} & \textbf{90.75} \\
			\hline
		\end{tabular}
	\end{table}
	
	\noindent Three findings stand out. First, FDM noise sensitivity escalates super-linearly with dimension: total accuracy degradation from $\sigma = 0$ to $\sigma = 0.20$ grows from 10.6\% (1D) to 27.7\% (2D) to 61.9\% (3D), while PINN degradation scales gracefully at 1.7\%, 3.7\%, and 7.6\% respectively. Second, at maximum noise in 3D, FDM accuracy collapses to 36\%, rendering it effectively unusable for practical thermal analysis, while PINN sustain approximately 91\% accuracy, a 2.5 times advantage. Third, the per-increment degradation rates recorded in Table~\ref{tab:noise_robustness} confirm that FDM accuracy in 3D declines monotonically and steeply across every noise level, while PINN accuracy remains comparatively stable, confirming that the performance gap is not a threshold effect but a structural difference in how each solver responds to boundary noise.
	
	\subsection{PINN Training Time Stability Across Noise Levels}
	
	\noindent Table~\ref{tab:pinn_noise_times} reports PINN training times across all 15 noise experiments. Training duration remains stable across noise levels within each dimension, with less than 2\% variation in 1D, 0.5\% in 2D, and 0.7\% in 3D, confirming that boundary noise magnitude does not influence optimization cost.
	
	\begin{table}[htbp]
		\centering
		\caption{PINN training times (seconds) across noise levels and dimensions.}
		\label{tab:pinn_noise_times}
		\begin{tabular}{lcccccc}
			\hline
			Dimension & $\sigma=0.00$ & $\sigma=0.05$ & $\sigma=0.10$ & $\sigma=0.15$ & $\sigma=0.20$ & Variation \\
			\hline
			1D & 68.54 & 67.71 & 68.33 & 67.97 & 67.28 & $<$2\% \\
			2D & 221.89 & 219.81 & 217.91 & 221.00 & 218.93 & $<$2\% \\
			3D & 894.19 & 892.97 & 892.78 & 892.73 & 892.97 & $<$1\% \\
			\hline
		\end{tabular}
	\end{table}
	
	\noindent \textit{Note: Execution times exhibit minor run-to-run variation ($<$5\%) attributable to shared GPU infrastructure scheduling; reported values are representative of a single consistent experimental run.}
	
	\subsection{Case Study: Boundary Temperature Reconstruction in Physical Units}
	\label{sec:case_study}
	
	\noindent To establish physical significance beyond relative error metrics, the 3D results were interpreted under the non-dimensionalization defined in Section~\ref{sec:methodology}, assigning $T_{\text{ref}} = 150$ K consistent with boundary temperatures reported for electronic chip thermal management \cite{cai}. Under $\sigma = 0.20$, corresponding to 30 K boundary measurement uncertainty, we evaluate mean absolute temperature reconstruction errors at the boundary face ($z = 0$ m) at $t = 60$ s, using the clean FDM solution as ground truth (Table~\ref{tab:case_study}).
	
	\begin{table}[htbp]
		\centering
		\caption{Mean absolute boundary temperature reconstruction errors in physical units.}
		\label{tab:case_study}
		\begin{tabular}{lcc}
			\hline
			Method & Mean Absolute Error (K) & Max Absolute Error (K) \\
			\hline
			Explicit FDM & 7.0 & 7.0 \\
			PINN & 2.1 & 6.8 \\
			\hline
			Reduction & 3.3$\times$ & $\dagger$ \\
			\hline
		\end{tabular}
	\end{table}
	
	\noindent FDM enforces the noisy boundary value as a hard constraint, propagating the full 30 K perturbation uniformly across the boundary face, a mean error of 7.0 K at a surface that should register 0 K relative to ambient. PINN, through soft boundary enforcement and physics-based regularization, suppresses this to a mean of 2.1 K, a 3.3 times reduction. The uniform FDM error (mean = max = 7.0 K) reflects the frozen noise protocol: a single noise trajectory is injected identically at all boundary nodes at each time step, with no smoothing applied. For thermal design applications where boundary temperature accuracy determines heat flux calculations and safety margins, a 4.9 K mean error difference is physically consequential.
	
	\noindent $\dagger$ Max errors are comparable (7.0 K vs 6.8 K) due to localized PINN boundary spikes; mean error is the appropriate metric for boundary reconstruction accuracy.
	
	\subsection{Computational Efficiency}
	
	\noindent Table~\ref{tab:computational_efficiency} summarizes resource requirements across methods. FDM offers dramatic speed advantages: four to five orders of magnitude faster than PINN training, with negligible memory footprint ($< 0.03$ MB) that remains dimension-independent. PINN memory consumption scales from 67 MB (1D, 0.4\% of GPU capacity) to 4,503 MB (3D, 27.5\%), indicating substantial headroom for larger-scale problems.
	
	\noindent The discretization efficiency ratio reveals a critical crossover: in 1D and 2D, PINN requires 4 times and 2.67 times more collocation points than FDM spacetime nodes respectively, offering no efficiency advantage. In these lower-dimensional regimes, the computational overhead of PINN training is unambiguously unjustified under clean conditions, and FDM should be the default choice. This reverses in 3D, where PINN requires 2.81 times fewer nodes than FDM while simultaneously delivering superior accuracy and noise robustness, the point at which the computational investment in PINN training begins to return structural dividends.
	
	\begin{table}[H]
		\centering
		\caption{Computational resource requirements and discretization efficiency across methods and dimensions.}
		\label{tab:computational_efficiency}
		\begin{tabular}{llccc}
			\hline
			Dim & Method & Memory (MB) & Spacetime Nodes & Efficiency Ratio \\
			\hline
			\multirow{2}{*}{1D}
			& Explicit FDM  & $< 0.03$ & 1,500 & \multirow{2}{*}{0.25$\times$} \\
			& PINN & 67 & 6,000   & \\
			\hline
			\multirow{2}{*}{2D}
			& Explicit FDM & $< 0.03$ & 22,500 & \multirow{2}{*}{0.38$\times$} \\
			& PINN & 919 & 60,000 & \\
			\hline
			\multirow{2}{*}{3D}
			& Explicit FDM & $< 0.03$ & 337,500 & \multirow{2}{*}{2.81$\times$} \\
			& PINN & 4,503 & 120,000 & \\
			\hline
		\end{tabular}
	\end{table}
	
	\subsection{Statistical Significance of Loss Weighting Strategy}
	
	\noindent Across 10 independent random seeds at $\sigma = 0.10$, adaptive gradient normalization achieves mean relative $L_2$ errors of 1.30\% (1D), 2.83\% (2D), and 7.26\% (3D), compared to 15.66\%, 16.44\%, and 40.80\% for equal weighting, improvements of 12 times, 5.8 times, and 5.6 times respectively (Table~\ref{tab:statistical_validation}). Mann-Whitney U tests reject the null hypothesis at $p = 9.13 \times 10^{-5} (< 0.0001)$ across all dimensions.
	
	\begin{table}[htbp]
		\centering
		\caption{Statistical comparison of loss weighting strategies.}
		\label{tab:statistical_validation}
		\begin{tabular}{llcccc}
			\hline
			Dimension & Weighting & Mean Rel.\ $L_2$ (\%) & Std Dev (\%) & CV (\%) & $p$-value \\
			\hline
			\multirow{2}{*}{1D}
			& Adaptive GradNorm & 1.304 & 0.453 & 34.73 & \multirow{2}{*}{$9.13 \times 10^{-5}$} \\
			& Equal & 15.661 & 7.367 & 47.04 & \\
			\hline
			\multirow{2}{*}{2D}
			& Adaptive GradNorm & 2.828 & 1.532 & 54.17 & \multirow{2}{*}{$9.13 \times 10^{-5}$} \\
			& Equal & 16.444 & 4.817 & 29.29 & \\
			\hline
			\multirow{2}{*}{3D}
			& Adaptive GradNorm & 7.263 & 2.647 & 36.45 & \multirow{2}{*}{$9.13 \times 10^{-5}$} \\
			& Equal & 40.800 & 14.524 & 35.60 & \\
			\hline
		\end{tabular}
	\end{table}
	
	\noindent Coefficients of variation remain acceptable across both strategies (29--55\%), confirming stable performance despite stochastic training dynamics. The consistent $p$-value across dimensions reflects the Mann-Whitney U test saturating at maximum discriminability given 10 samples per group--the practical significance is better conveyed by the error magnitude ratios than the $p$-value alone.
	
\begin{figure}[H]
		\centering
		\includegraphics[scale=0.65]{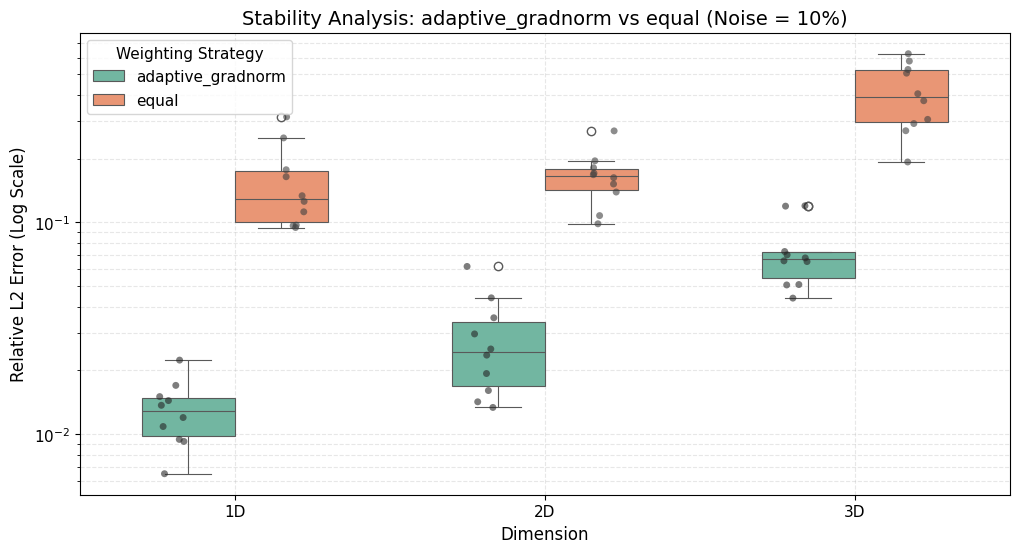}
		\caption{Distribution of relative $L_2$ errors across 10 independent seeds for adaptive gradient normalization (teal) and equal weighting (orange) strategies under $\sigma = 0.10$ noise, shown for each dimension. Box plots illustrate median, interquartile range, and outliers.}
		\label{fig:statistical_test}
	\end{figure}

	\noindent Figure~\ref{fig:statistical_test} displays the error distributions as box plots on a log scale. The separation between adaptive gradient normalization and equal weighting is visually unambiguous across all dimensions, with no overlap between interquartile ranges in any dimension. The increasing spread of equal weighting from 1D to 3D further corroborates the worsening instability of equal loss balancing with dimensionality.
	
	\section{Discussion}
	\label{sec:discussion}
	
	\noindent The results establish distinct performance regimes that together define when each solver class is appropriate. We examine the underlying mechanisms, contextualise findings against the literature, and derive practical guidance for method selection.
	
	\subsection{Accuracy and Computational Efficiency Trade-offs}
	
	\noindent Under clean boundary condition, PINN achieves 2.72 times, 1.43 times, and 1.04 times lower relative $L_2$ errors than optimal FDM schemes across 1D, 2D, and 3D respectively. The diminishing margin is mechanistically expected: as dimensionality increases, solution complexity and FDM discretization error both grow, compressing the accuracy gap until both methods converge toward similar floors. For sufficiently high-dimensional clean problems, computational cost becomes the primary differentiator. The four-to-five orders of magnitude speed advantage of Explicit FDM over PINN training confirms the assessments of Sharimbayev et al. and Grossmann et al. for Poisson and elliptic equations \cite{batyr, grossmann}. For the single forward problem studied here, FDM's speed advantage is unambiguous. For parametric studies varying geometry or boundary condition, this calculus shifts in PINN's favor, consistent with Cai et al.'s optimization demonstrations \cite{cai}, as the trained network evaluates arbitrary spatiotemporal coordinates in milliseconds at negligible incremental cost. The $L_\infty$ anomaly in 3D, where PINN yields 0.0866 versus FDM's 0.0622 despite superior $L_2$ performance (Table~\ref{tab:clean_accuracy}), arises from localized errors at the eight domain corners where three boundary conditions intersect, creating gradient discontinuities. This reflects a single trade-off: PINNs minimize global error through a continuous approximation, while FDM distributes error more uniformly via local stencils. Consequently, applications governed by maximum pointwise error (e.g., thermal stress hotspots) favor FDM, whereas global reconstruction tasks favor PINN. Figures~\ref{fig:1d_results}, \ref{fig:2d_results}, and \ref{fig:3d_results} show that FDM enforces near-zero boundary error, while PINN retains small residuals due to soft enforcement through $\mathcal{L}_{\text{BC}}$ \cite{leveque, strikwerda, raissi, lulu}. Under clean conditions this is a minor limitation, but under noise it becomes advantageous by preventing direct propagation of corrupted boundary values.
	
	\subsection{Physical Mechanisms of Noise Robustness}
	
	\noindent The divergent noise sensitivity between FDM and PINNs follows directly from boundary treatment. FDM imposes boundary values through direct assignment, so noise injected at the boundaries propagates deterministically into the domain. In three dimensions, this effect is amplified geometrically: faces introduce independent noise, edges accumulate multiple contributions, and corners concentrate three-way interactions, producing the super-linear degradation observed in Table~\ref{tab:noise_robustness}: 10.6\% (1D) $\to$ 27.7\% (2D) $\to$ 61.9\% (3D). PINNs mitigate this through multi-objective optimization. Noisy boundary conditions enter as soft constraints in $\mathcal{L}_{\text{BC}}$, balanced against $\mathcal{L}_{\text{PDE}}$ and the clean $\mathcal{L}_{\text{IC}}$. The PDE residual acts as a physics-based regularizer, while the IC loss anchors temporal consistency, yielding implicit denoising of the solution. Adaptive gradient normalization is essential: without it, boundary loss dominates early training and drives overfitting to noise. Dynamic reweighting maintains balance across objectives, producing the observed 12 times, 5.8 times, and 5.6 times error reductions over equal weighting ($p < 9.13 \times 10^{-5}$). Two complementary mechanisms mitigate the reward-hacking and physics loss divergence failure modes identified by Jekic et al. \cite{jekic}: the frozen noise protocol preserves temporal consistency across training epochs and dimension-specific weight clamping prevents physics loss from being overwhelmed. Collectively, these observations establish that PINN noise robustness is not an inherent property but an emergent behavior arising from proper architectural choices, loss balancing strategies, and problem formulation.
	
	\subsection{Dimensional Scaling and Discretization Efficiency}
	
	\noindent The discretization efficiency crossover at 3D emerges from fundamentally different scaling laws. FDM requires uniform grid coverage scaling as $\mathcal{O}(N^d \times T)$: the 3D configuration demands $15^3 \times 100 = 337{,}500$ nodes. Implicit and Crank-Nicolson schemes additionally solve sparse linear systems at each time step, with interior node counts scaling from 13 (1D) to 169 (2D) to 2,197 (3D), a disparity that grows cubically with dimensionality. Grossmann et al. \cite{grossmann} demonstrated that PINNs scale more favorably with dimensionality than grid-based methods through automatic differentiation eliminating the curse of dimensionality. The present results provide the first systematic empirical confirmation for heat diffusion specifically, identifying the crossover between 2D and 3D. Table~\ref{tab:pinn_noise_times} confirms that PINN training time remains stable across all noise levels within each dimension. This is distinct from the theoretical result of André-Sloan et al. \cite{sebastien}, whose bound $d_N \log d_N \gtrsim N_s \eta^2$ concerns model capacity scaling, not training duration. Stability follows from the composition of the collocation set: noisy boundary targets comprise only $N_{\text{BC}}$ points among 6,000--120,000 total samples, while PDE and IC losses computed on clean physics dominate the training signal.
	
	\subsection{Comparative Performance in Context}
	
	\noindent Bueno et al.'s finding that PINN outperforms FTCS and ADI for 2D heat equations \cite{vitor} is confirmed here; the 3D extension reveals this advantage does not uniformly scale, with problem complexity in higher dimensions offsetting PINN's temporal advantage under clean conditions. The noise robustness findings in this work are consistent in direction with prior works but exceed previous quantitative assessments in magnitude. Wong et al. \cite{jian} demonstrated a 10 times effective noise reduction for PINN versus data-driven networks on Navier-Stokes; the present work demonstrates a 2.5 times accuracy advantage over physics-based FDM under maximum noise in 3D, a more stringent comparison. Unlike B-PINN \cite{yang} and naPINN \cite{hankyeol}, which achieve robustness through Bayesian weight inference and energy-based noise modeling respectively, comparable structural resilience is demonstrated here through loss balancing alone, without architectural modification or assumed noise distributions. Bajaj et al.'s GP preprocessing \cite{bajaj} confirms that boundary noise suppression is achievable in forward PINN settings; the present work extends this to 3D and quantifies dimensional dependence without a preprocessing stage. The fPINN results of Pang et al. \cite{guofei} corroborate that physics-based regularization confers noise resilience across PDE classes; the present findings establish that this resilience degrades gracefully with dimensionality rather than collapsing, previously unquantified for heat diffusion. Across all studies, physics-based regularization provides substantial noise resilience; the present contribution is the first systematic quantification of how this resilience scales from 1D to 3D under identical frozen noise protocols. The failure of Adam$\to$L-BFGS switching (Table~\ref{tab:best_pinn_configs}) contrasts with common PINN practice \cite{grossmann, jekic, batyr}, likely stemming from collocation resampling every 100 epochs preventing the quasi-Newton approximation from accumulating accurate curvature information.

	\section{Conclusion}
	\label{sec:conclusion}
	
	\noindent Transient heat diffusion under noisy boundary condition exposes a structural vulnerability in classical grid-based solvers that worsens with dimensionality. This work establishes a systematic quantification of this vulnerability across three spatial dimensions, identifying the precise regimes where the classical solver paradigm is insufficient and PINN becomes the methodologically justified choice. Under clean boundary condition, PINN achieves superior relative $L_2$ accuracy across all dimensions, with error reductions of 2.72 times (1D), 1.43 times (2D), and 1.04 times (3D) over base FDM schemes. The decisive separation, however, emerges under noise. At maximum boundary noise ($\sigma = 0.20$) in three dimensions, FDM accuracy collapses to 36\% while PINN sustains approximately 91\%, a 2.5 times advantage. This divergence is mechanistically grounded: FDM enforces boundary condition as hard assignments, propagating and geometrically amplifying noise through repeated stencil applications, while PINN treats noisy boundaries as soft constraints balanced against physics residuals and clean initial conditions. A physical case study confirms dimensional significance: under 30 K boundary noise on a 150 K copper thermal system, FDM produces a mean boundary reconstruction error of 7.0 K versus 2.1 K for PINN, a 3.3 times reduction with direct implications for thermal design margin calculations. Furthermore, a critical discretization efficiency crossover emerges at 3D, where PINN require 2.81 times fewer spacetime nodes than FDM while achieving superior accuracy and noise tolerance.
	
	\noindent These findings carry direct operational consequences for solver selection. For real-time, low-dimensional applications with precisely known boundary condition, FDM remains the unambiguous choice: millisecond solve times, negligible memory requirements, and well-understood stability criteria make it deployable on resource-constrained hardware. The case for PINN, however, becomes necessary rather than merely preferable at the threshold $\sigma \geq 0.15$ in three dimensions, where FDM accuracy falls to approximately 52\% while PINN maintains 93\%. The $\approx$893 seconds training investment is justified in these contexts, as the trained network amortizes across unlimited subsequent queries at negligible incremental cost. A critical architectural finding accompanies this regime boundary: adaptive gradient normalization is not optional. It delivers 5.6 to 12 times error reduction over equal weighting and must be treated as a structural requirement for any PINN deployment under noisy boundary condition, not a tunable preference.
	
	\noindent A boundary condition on the scope of these findings warrants explicit statement. The present study employs homogeneous Dirichlet boundary condition, where noise magnitude $\sigma$ is expressed relative to the characteristic temperature scale under the non-dimensionalization $\theta = T/T_{\text{ref}}$, not relative to the boundary value itself. The operational thresholds identified here ($\sigma \geq 0.15$ for PINN necessity) and the physical errors reported in the copper case study should therefore be interpreted relative to solution magnitude. Extending the framework to non-homogeneous boundary condition, where noise can be rigorously defined relative to the boundary value, represents the natural next step toward physically grounded thresholds for industrial thermal systems. Further generalization to anisotropic media, temperature-dependent thermal conductivity, and stochastic boundary processes would broaden the operational regimes established here into the full complexity of real engineering environments further strengthening the case for PINNs as the solver of choice under noise and dimensionality.

	\section*{Code Availability}
	The source code for all experiments, including FDM implementations, PINN training scripts, hyperparameter search, and noise robustness analysis, is publicly available at \href{https://github.com/shreeshb51/overcoming_the_limits_of_fdm_pinn_for_noisy_high_dimensional_heat_diffusion}{GitHub}.

\end{document}